\documentclass[sigconf]{acmart}

\AtBeginDocument{%
  \providecommand\BibTeX{{%
    \normalfont B\kern-0.5em{\scshape i\kern-0.25em b}\kern-0.8em\TeX}}}

\setcopyright{acmlicensed}
\copyrightyear{2025}
\acmYear{2025}
\acmDOI{XXXXXXX.XXXXXXX}

\usepackage{multirow}

\usepackage{amssymb}
\usepackage{bbding}
\usepackage{colortbl}
\usepackage{xcolor}
\usepackage{stfloats}

\usepackage{hyperref}

\hypersetup{
    colorlinks=true,
    linkcolor=blue,
    citecolor=blue,
    urlcolor=blue
}

\begin{document}

\title{Repeating Words for Video-Language Retrieval with Coarse-to-Fine Objectives}

% \author{Haoyu Zhao}
% \affiliation{
%   \institution{Fudan University}}

% \author{Jiaxi Gu}
% \affiliation{
%   \institution{Tencent}}

% \author{Shicong Wang}
% \affiliation{
%   \institution{Fudan University}}

% \author{Xing Zhang}
% \affiliation{
%   \institution{Fudan University}}

% \author{Hang Xu}
% \affiliation{
%   \institution{Huawei Noah’s Ark Lab}}

% \author{Zuxuan Wu}
% \affiliation{
%   \institution{Fudan University}}

% \author{Yu-Gang Jiang}
% \affiliation{
%   \institution{Fudan University}}

\author{Haoyu Zhao$^1$, Jiaxi Gu$^2$, Shicong Wang$^1$, Xing Zhang$^1$, Hang Xu$^3$, \\ Zuxuan Wu$^1$, Yu-Gang Jiang$^1$}
 
\affiliation{%
 \institution{$^1$Fudan University, $^2$Tencent, $^3$Huawei Noah’s Ark Lab}
 \country{}
}

\renewcommand{\shortauthors}{H. Zhao, et al.}

\begin{abstract}
The explosive growth of video streaming presents challenges in achieving high accuracy and low training costs for video-language retrieval. However, existing methods rely on large-scale pre-training to improve video retrieval performance, resulting in significant computational demands. Additionally, the fine-grained information in videos and texts remains underexplored.
To alleviate these problems, we propose a novel framework to learn fine-grained features for better alignment and introduce an inference pipeline to improve performance without additional training.
Specifically, we employ coarse-to-fine objectives to understand the semantic information of video-text pairs, including contrastive and matching learning. The fine-grained data used for training is obtained through the Granularity-Aware Representation module, which is designed based on similarity analysis between video frames and words in captions.
Furthermore, we observe that the repetition of keywords in the original captions, referred to as ``Repetition'', can enhance retrieval performance and improve alignment between video and text. Based on this insight, we propose a novel and effective inference pipeline that incorporates a voting mechanism and a new Matching Entropy metric to achieve better retrieval performance without requiring additional pre-training.
Experimental results on four benchmarks demonstrate that the proposed method outperforms previous approaches. Additionally, our inference pipeline achieves significant performance improvements, with a 2.1\% increase in Recall@1 on the MSR-VTT dataset and a 1.6\% increase on the DiDeMo dataset.
\end{abstract}

\keywords{Video retrieval, Coarse-to-fine objectives, Repeating words}

\maketitle

\section{Introduction}
Video-Language Retrieval (VLR) task focuses on accurately retrieving the corresponding video with the provided caption or the corresponding caption with the given video data. Existing methods~\cite{Shvetsova2022everything, Croitoru2021teachtext, Dzabraev2021Mdmmt}, which employ transformer-based structure, have achieved good performance on this task.
This technology is applied in many real-world scenarios, including television shows, digital libraries, and online videos.
Besides, methods~\cite{Luo2022CLIP4Clip, Xu2021VideoCLIP} with pre-training paradigm also select the VLR as the downstream task.

\begin{figure}[t]
\begin{center}
\includegraphics[width=1.0\linewidth]{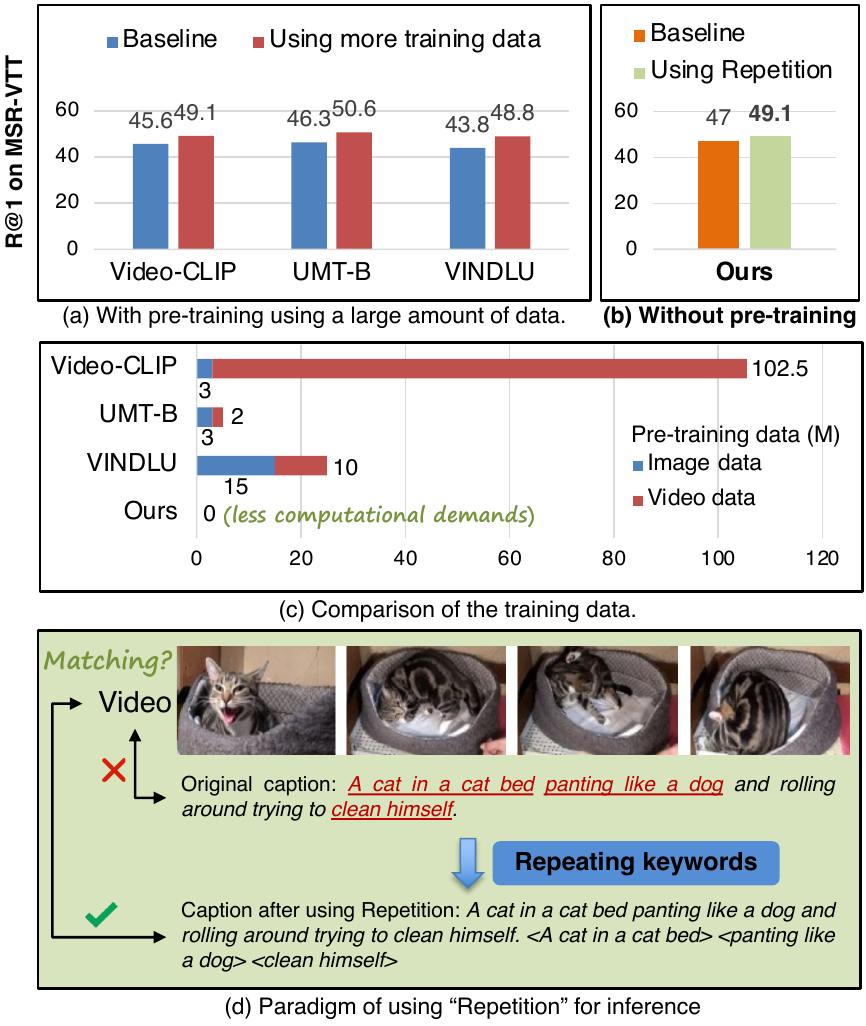}
\end{center}
   \caption{Comparison to state-of-the-art methods, \textit{i.e.}, Video-CLIP~\cite{Xu2021VideoCLIP}, UMT-B~\cite{li2023umt-b}, and VINDLU~\cite{cheng2023vindlu}, in text-to-video retrieval performance on MSR-VTT dataset. In (a), the blue bars indicate performance with less training data, while the red bars show the improvements achieved after pre-training with more data. The amount of data used is shown in (c). In (b), we demonstrate that our method can achieve significant improvements without pre-training by solely utilizing the ``Repetition'' strategy. We also give an example of the ``Repetition'' in (d), which denotes adding keywords to the original caption for better multimodal retrieval.}
\label{fig: performance and data}
\vspace{-5pt}
\end{figure}

However, there are two challenges that exist in the video retrieval domain. 
On the one hand, previous methods~\cite{Liu2022Hierarchical, Hendricks2018Localizing, Mithun2018Learning, Wray2019Fine, Zhu2020ActBERT, Lei2021Less} use a coarse-grained alignment for video and caption data, failing to fully exploit the potential of fine-grained clues.
Although existing methods~\cite{chen2020fine, wang2023unified, wang2022align, ma2022x-clip} employ the fine-grained information from video and caption to enhance the accuracy of video retrieval, these approaches lack an efficient way to introduce the fine-grained information to the training stage.
For instance, the UCOFIA model~\cite{wang2023unified} employs the Sinkhorn-Knopp algorithm to calculate the similarity score of each granularity, which only assists in calculating the retrieval matching score instead of enhancing the understanding of data.
Although methods~\cite{wang2022align, ma2022x-clip} attempt to use contrastive learning loss between fine-grained video frames and words to enhance performance, the model's alignment and performance are still limited.

On the other hand, existing state-of-the-art methods~\cite{Xu2021VideoCLIP, li2023umt-b, cheng2023vindlu} deeply rely on large parameters and large-scale pre-training data to improve the performance on downstream tasks, including the VLR. In Fig.~\ref{fig: performance and data} (a) and (b), we compare the text-to-video performance of models Video-CLIP~\cite{Xu2021VideoCLIP}, UMT-B~\cite{li2023umt-b}, VINDLU~\cite{cheng2023vindlu}, and ours on the MSR-VTT~\cite{xu2016msrvtt} dataset.
The compared methods are all following the ``pre-training'' paradigm.
When maintaining the structure of the model, we find that these methods all show improved performance on the downstream retrieval task with larger pre-training data.
\textit{However, the limitations of these methods are also clear, such a paradigm requires more computational resources and increases the model's dependency on large-scale training data.}

In addition, without using more pre-training data, it seems hard to further improve the model's performance. Fortunately, we find inspiration from the human language learning process.
In the past several decades, \textit{language repetition} has received considerable attention from anthropologists, linguists, and educators~\cite{moore2011language}. Repetition means repeating words or sentences during speaking and writing. It is important to repeat words in the language learning process. Literature~\cite{duff2000repetition} indicates that repetition in conversation can serve to consolidate what is being learned and theory~\cite{harmon2021theory} analyzes that repeating words helps access new contents by reactivating preceding words in language production. Moreover, language repetition task~\cite{klem2015sentence-repetition, mcdade1982use-repetition, rodd1971children-repetition, schwartz1978elicited-repetition, alloway2005role-repetition} is also used for evaluation of perception, vocabulary knowledge, grammatical processing, and speech production. 
It can be seen that the repetition of words can help humans to learn the language. 
\textit{Because the VLR models also attempt to comprehend natural language, we are motivated by these findings to explore repeating words in captions within the VLR framework.}

In this paper, we propose a novel coarse-to-fine VLR framework without using pre-training data to achieve good performance. Specifically, during the training stage, the framework contains two coarse-grained objectives and one fine-grained objective, \textit{i.e.,} the Video-Caption Contrastive, the Video-Caption Matching, and the Frame-Title Matching. 
These objectives ensure our model focuses on both the coarse-grained video-caption data and the fine-grained frame-title data, introducing fine-grained details at the training stage.
We also have an important insight that the matching objective is more suitable than the contrastive objective for fine-grained information. It can help the model better understand the detailed feature representation.
To obtain the frame and title-level data, we introduce the Granularity-Aware Representation module, which can generate representative frames and concise keywords. 

Furthermore, we first propose a simple but effective ``Repetition'' strategy in our VLR framework, shown in Fig~\ref{fig: performance and data} (d). 
Our observation indicates that the model achieves better alignment in certain video-caption instances when the keywords are intentionally repeated in the caption.
To precisely identify these instances, we develop an inference pipeline that implements a voting mechanism and introduces a Matching Entropy metric to assess voter consistency.
We repeat the keywords to the captions for those instances that exhibit lower ME values.
We also find that indiscriminately repeating words in every case is ineffective.
Experimental results on four benchmarks demonstrate the superior performances of our proposed framework with coarse-to-fine objectives and the ``Repetition'' strategy can enhance the model's performance without increasing the training data.

The main contributions are summarized as follows.

\begin{itemize}
\item We propose a novel coarse-to-fine framework for the VLR task, along with the Granularity-Aware Representation module to capture fine-grained information. The framework consists of two coarse-grained objectives and one fine-grained matching objective, which enables better alignment between video and caption. We find that the matching objective is more suitable than the contrastive one for fine-grained data.
\item We propose a video-language retrieval inference pipeline with the ``Repetition'' strategy to improve the model's performance during the inference stage, without requiring extra computationally intensive pre-training. In our inference pipeline, we include a voting mechanism and the proposed Matching Entropy metric to select the appropriate candidates for applying the ``Repetition'' strategy.
\item Experimental results on MSR-VTT, DiDeMo, MSVD, and YouCook2 benchmarks demonstrate the superior performances of our framework. The proposed ``Repetition'' strategy also provides a novel perspective for the retrieval model.
\end{itemize}

\section{Related Work}

\subsection{Video-Language Retrieval}

Most of the previous works for the VLR task use multi-modal encoders to align the vision and language representation. Early methods design intensive fusion structures based on CNN for cross-modal learning, such as~\cite{Liu2022Hierarchical, Hendricks2018Localizing, Mithun2018Learning}. With the excellent performance of ViT model~\cite{Dosovitskiy2020An}, the existing methods~\cite{Cheng2021Improving, Devlin2019BERT} are mainly based on transformer structure. Approaches~\cite{Zhu2020ActBERT, Lei2021Less} adopt Bert model~\cite{Devlin2019BERT} as language encoder to get a better understanding of text information. Yang et al.~\cite{Yang2021Taco} proposes a token-aware cascade model to improve contrastive learning performance. Methods~\cite{Liu2021Progressive, Lin2022Text} analyze the modality gap issue and focus on the vision-language alignment. 
With the popularity of the “Pre-training” and “fine-tune” paradigm, the pre-training model~\cite{Radford2021Learning, Li2022Blip} is widely used in many fields, including the retrieval task. Works~\cite{Zhuo2022CLIP4Hashing, Zhang2023Multimodal} transfer the image pre-training model to the video-language retrieval method. Bridging the large semantic gap across different modalities is a research challenge, as recent works~\cite{Ge2022MILES, Wang2022Boosting, Liu2021HiT} have demonstrated. Ge et al.~\cite{Ge2022Bridgeformer} proposes a bridge model to do video-text interactions and keeps the high efficiency for retrieval. 
Although the above methods achieve good results, they align the video language with a coarse-grained global matching, not fully addressing the fine-grained clues. Additionally, the pre-training manner also makes models rely on large-scale training data, which brings heavy computation costs.

\subsection{Fine-grained Data for Video Retrieval}
With the development of deep learning~\cite{Zhu2020ActBERT, Lei2021Less, tang2023prototransfer,zhao2023scene, Wray2019Fine, zhao2021space, zhao2024magdiff, zhao2022need, zhao2023memory}, the coarse-grained and fine-grained data are also popular for learning the video representation and video-language understanding.
Methods~\cite{Liu2022FeatInter, Wang2020Learning} consider the object-level information from video via object detection approaches. The objects in the video contain more fine-grained features than the whole video. Concurrent to our work, method~\cite{wang2023unified, ma2022x-clip, Wang2022Alignandtell} select the frames in the video and the words in the caption as the fine-grained data. UCOFIA model~\cite{wang2023unified} employs the frame data to calculate the retrieval matching score. X-CLIP model~\cite{ma2022x-clip} and Align\&tell model~\cite{Wang2022Alignandtell} utilize the high-dimension features of fine-grained data and design the contrastive loss to extract the related features. However, because these models use lots of negative samples to learn discrimination when aligning fine-grained visual-textual data, they cannot directly match video and text features during training, resulting in limited performance. Therefore, it is necessary to make better use of fine-grained data.

\begin{figure*}[t]
    \centering
    \includegraphics[width=1.0\linewidth]{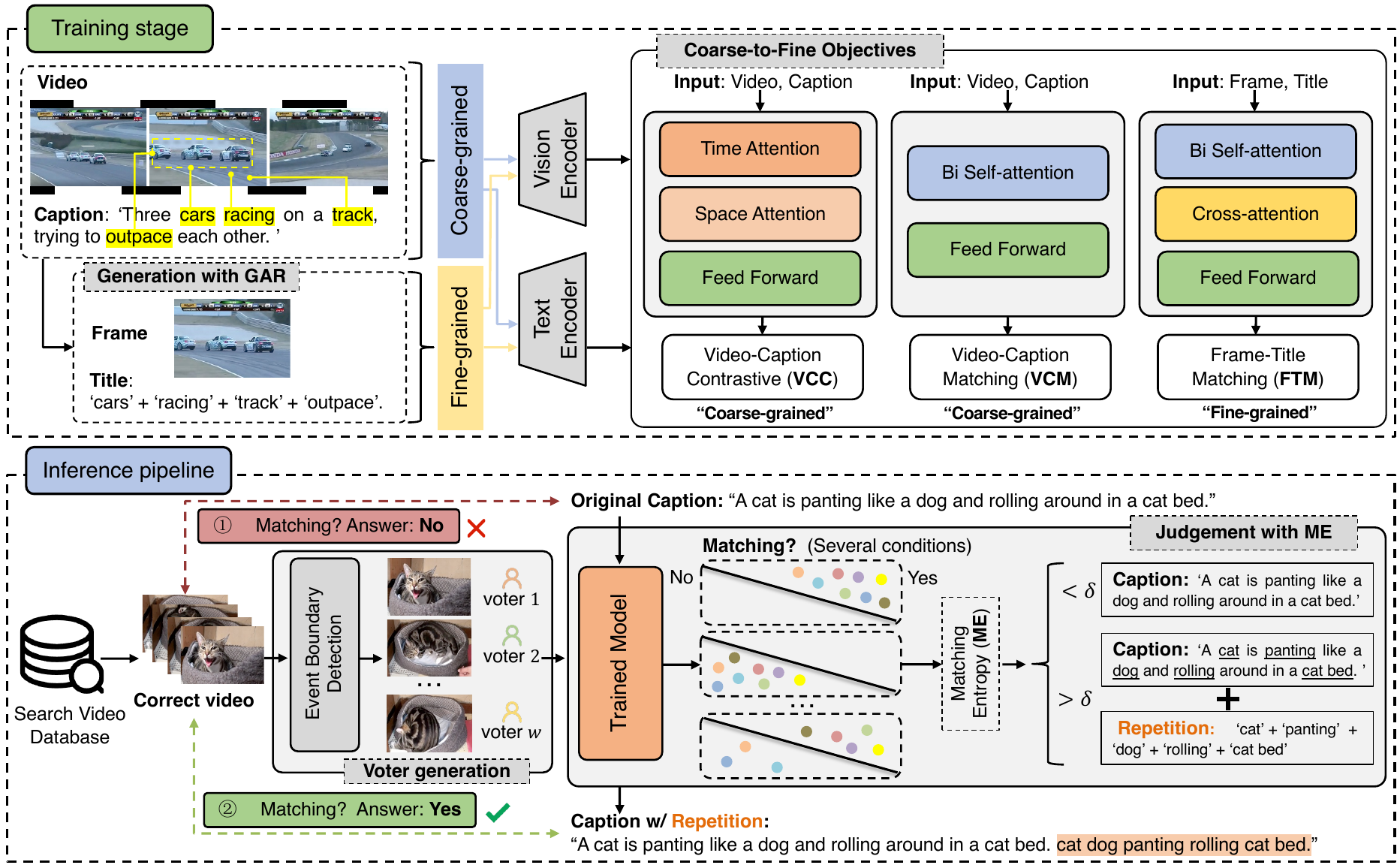}
    \caption{Overview of the proposed video-language retrieval framework. During the training stage, we carefully select three types of coarse-to-fine objectives to understand video content and establish connections between semantic information of video and language. In this stage, fine-grained information is extracted using the Granularity-Aware Representation module. To enhance retrieval performance while avoiding significant training costs, we develop a novel inference pipeline incorporating the proposed ``Repetition'' strategy.}
    \label{fig:pipeline}
    \vspace{-5pt}
\end{figure*}

\subsection{Repeating Language Words}
In the human language learning process, repeating words is regarded as very important~\cite{moore2011language}.
In previous research, Literatures~\cite{klem2015sentence-repetition, mcdade1982use-repetition, rodd1971children-repetition, schwartz1978elicited-repetition, alloway2005role-repetition} demonstrate that the repetition of sentences and words is a useful indicator for measuring individual differences in language ability within human language learning.
Literature~\cite{moore2011language} represents that repetition is a resource that can be used to do many different things, including language socialization.
The repetition is also employed in language classroom interaction~\cite{duff2000repetition} and retrieval in language production~\cite{harmon2021theory}.
In conclusion, these works and findings suggest that repeating words mechanism plays an important role in the language learning process. 
Since VLR models also aim to understand natural language sentences or words, it's both feasible and important to investigate the use of repeating words in this field.

\section{Methods}

In this work, we propose a novel framework to achieve video-language retrieval task, shown in Fig.~\ref{fig:pipeline}.
To better utilize the fine-grained information of video-caption data, we propose three coarse-to-fine objectives for joint model training, including two coarse-grained objectives and one fine-grained objective, \textit{i.e.,} the Video-Caption Contrastive (VCC), the Video-Caption Matching (VCM), and the Frame-Title Matching (FTM). To obtain the fine-grained information from the origin data-caption data, we introduce the Granularity-Aware Representation (GAR) module to get a representative frame and generate a concise title that captures the main content of the video. 
Furthermore, to improve the model's performance without using extra training data, we propose a novel ``Repetition'' strategy and an inference pipeline. ``Repetition'' denotes the noun and verb words in the caption.
The method utilizes the voting mechanism and the Matching Entropy metric to select suitable cases to use the strategy.

\subsection{Granularity-Aware Representation Module}

\begin{figure}[t!]
\centering
\includegraphics[width=0.85\linewidth]{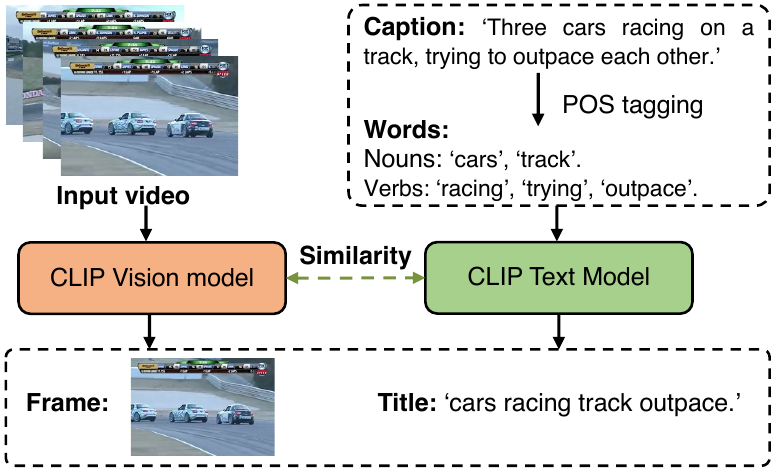}
   \caption{Overview of the proposed Granularity-Aware Representation module. The abundant information contained in video media makes video content retrieval more challenging. We aim to extract fine-grained information, such as frames and keywords, to help the model better understand representative features.}
\label{fig:gar}
\vspace{-5pt}
\end{figure}

Existing fine-grained works~\cite{wang2023unified, Wang2022Alignandtell} for retrieval employ all the words in the caption and all the patch features from the video as the fine-grained data, which also bring irrelevant information into the process of handling fine-grained features. Although method~\cite{ma2022x-clip} selects fine-grained words and frames instead of the whole caption, these data are from high-dimension, thereby losing the low-dimensional information for vision-text alignment. Inspired by these methods, to obtain fine-grained information from video-caption data, we propose the Granularity-Aware Representation (GAR) module, which can generate a recapitulative title and a representative frame for each video, as shown in Fig.~\ref{fig:gar}. The outputs are used for the coarse-to-fine objectives.

Firstly, we get the video $V$ and the corresponding captions $C$.
For each video, we select $m$ frames $\{F_{1}, F_{2}, F_{3}, ..., F_{m}\}$ uniformly.
We process the captions $C$ using the Part of Speech (PoS) tagging to extract the set of nouns and verbs, denoted as $\{n^1, n^2, n^3,...,n^p\}$ and $\{vb^1, vb^2, vb^3,...,vb^q\}$.
To quantify the semantic correlation between words and visual frames, we employ the CLIP model ~\cite{Radford2021Learning} ${\mathcal M}_{clip}$ for similarity calculation, which possesses a certain level of matching capability. For each noun $n^k$, we get its similarity ${\mathcal R}_{(n^k,F)}$ with $m$ frames, denoted as Eq.~\ref{Eq: 1}. Similarly, the $\mathcal {R}_{(vb^k,F)}$ of verb $vb^k$ is obtained in the same way.

\begin{equation}\label{Eq: 1}
\mathcal{R}_{(n^k,F)}=\sum_{i=1}^m \mathcal{M}_{c l i p}\left(n^k, F_i\right)
\end{equation}

Subsequently, based on the rank of calculated similarity scores, we can select the nouns and verbs and combine them as the new Title $\mathcal{T}$, which are the textual representations corresponding to the video's content.
For example, as Fig.~\ref{fig:gar} shows, the original caption is ``\textit{Three cars racing on a track, trying to outpace each other.}'' and the new Title is ``\textit{cars racing track outpace}''.
Because the Title is generated through similarity analysis from video-caption, it is fine-grained information derived from the original data, which helps eliminate irrelevant words.

In addition, besides the fine-grained Title $\mathcal{T}$, we also generate a frame-granularity representation from video, represented as $\mathcal{F}$.
Specifically, we select a single frame from each video that corresponds to its respective title, streamlining the frame selection process.
By calculating the similarity between each frame and the Title, the Frame $\mathcal{F}$ is got by ranking to get the top-1 frame. The Frame $\mathcal{F}$ not only aligns with the content of the title but also represents the content in the video. The process is denoted as:

\begin{equation}\label{Eq: 2}
\mathcal{F}=Rank({\mathcal M}_{clip}(\mathcal{T}, \{F_{1}, F_{2}, F_{3}, ..., F_{m}\}))
\end{equation}

Considering the characteristics of the $\mathcal{T}$ and $\mathcal{F}$, our coarse-to-fine objectives encompass a combination of video-caption contrastive, matching, and frame-title matching.

\subsection{Coarse-to-Fine Objectives}

Following BLIP model~\cite{Li2022Blip}, we optimize a video-language retrieval framework that contains three coarse-to-fine objectives, shown at the top part of Fig ~\ref{fig:pipeline}. 
Based on the BLIP model, we design our framework in a coarse-to-fine manner, which incorporates the coarse-grained data (video-caption) and fine-grained data (Title $\mathcal{T}$ - Frame $\mathcal{F}$).
We mainly adopt the coarse-grained contrastive and matching objectives, and the fine-grained matching objective to align the visual and textual information.

\subsubsection{Coarse-grained Objectives}
\ 
\newline
\textbf{Video-Caption Contrastive.}
Literature~\cite{Li2021Align} first introduces the contrastive objective between the image and text to learn the representation and the alignment.
The goal of contrastive is to align the visual feature and text feature by reducing the distance of positive pairs and repelling negative pairs. 
In the VLR task, it is reasonable to transfer such an objective into the alignment between the video and caption.
Following ~\cite{Li2021Align}, two queues are maintained to store $M$ video-caption pairs and the normalized video and caption features from the momentum encoders are denoted as $v_{m}$ and $c_{m}$, respectively. 
Then, the VCC objective is defined as $\mathcal{L}_{v 2 c}+\mathcal{L}_{c 2 v}$ as the follows:

\begin{equation}\label{Eq. 3}
\mathcal{L}_{v 2 c}\left(v_i\right)=-\sum_{k \in M} \log \frac{\exp \left(v_i^T c_k / \tau\right)}{\sum_{m=1}^M \exp \left(v_i^T c_m / \tau\right)}
\end{equation}

\begin{equation}\label{Eq. 4}
\mathcal{L}_{c 2 v}\left(v_i\right)=-\sum_{k \in M} \log \frac{\exp \left(c_i^T v_k / \tau\right)}{\sum_{m=1}^M \exp \left(c_i^T v_m / \tau\right)}
\end{equation}

\noindent \textbf{Video-Caption Matching.}
Matching objective~\cite{Li2022Blip} is a binary classification task that predicts the compatibility between vision and language. For the coarse-grained objective, we define the matching loss between the video and caption, denoted as VCM. We use a fully connected layer to predict the alignment probability $p_{vcm}$ followed by \textit{softmax} operation. We define the VCM as:

\begin{equation}\label{Eq. 5}
\mathcal{L}_{v c m}=\mathrm{E}_{(v, c) \sim D} \mathrm{H}\left(y_{v c m}, p_{v c m}(v, c)\right)
\end{equation}

where $y_{vcm}$ is the ground-truth. Since VCM is less dependent on negative samples, it can simplify the training process and potentially reduce the amount of training data required.

\subsubsection{Fine-grained Objective}
\ 
\newline
\textbf{Frame-Title Matching.} In the GAR module, we prepare the Title $\mathcal{T}$ and Frame $\mathcal{F}$ information as the fine-grained information. In the existing fine-grained VLR method X-CLIP~\cite{ma2022x-clip}, they define two contrastive objects for the both coarse-grained and fine-grained data. However, compared to the contrastive objective, the matching objective can align the relevant features in the video and caption more finely, rather than simply aligning the entire video with a whole piece of text. This helps to handle more complex tasks, such as detailed descriptions within a video or regional text correspondence. We also demonstrate this conclusion through the comparison with existing fine-grained methods in Table~\ref{tab: compare with other fine-grained method} and the ablation studies in Table~\ref{tab: evaluation of different objectives}.
So, we use the Frame-Title Matching (FTM) objective instead of the contrastive for fine-grained Title $\mathcal{T}$ and Frame $\mathcal{F}$. As shown in Fig.~\ref{fig:pipeline}, the Title is put into the bi-directional self-attention layer and the Frame is handled by the cross-attention layer. We define the FTM as:

\begin{equation}\label{Eq. 5}
\mathcal{L}_{f t m}=\mathrm{E}_{(f, t) \sim D} \mathrm{H}\left(y_{f t m}, p_{f t m}(f, t)\right)
\end{equation}

where $p_{ftm}$ is the alignment probability and the $y_{ftm}$ represents the ground-truth. Overall, we use the sum of VCC, VCM, and FTM as the total loss $\mathcal{L}$ to train the model, as follows:

\begin{equation}\label{Eq: total loss}
\mathcal{L} = \mathcal{L}_{v 2 c} + \mathcal{L}_{c 2 v} + \mathcal{L}_{v c m} + \mathcal{L}_{f t m}
\end{equation}

\subsection{Inference Pipeline with ``Repetition''}

In our work, we propose a novel ``Repetition'' strategy, whose key idea involves repeating words to the original caption, referred to as ``Repetition''.
We show an example in the bottom part of Fig.~\ref{fig:pipeline}. The original caption is ``\textit{A cat is panting like a dog and rolling around in a cat bed.}'' and we repeat the words ``\textit{cat, dog, panting, rolling, cat bed}'' to the original caption. 
The ``Repetition'' refers to the repeated use of noun and verb words in the caption.
We find that the model judges the original caption and correct video as not matching, but after repeating the words, the model makes the right answer.
Furthermore, we find that this strategy can effectively improve the performance of the model in the VLR task. 
However, we also find that not all the cases in the test set are suitable for this strategy. In Table~\ref{tab: impact of inference pipeline}, we demonstrate that simply adding the ``Repetition'' to all cases in the test set will even harm its performance.
To improve the model's performance on VLR test sets, we need to select suitable cases for our ``Repetition''. So, we design an inference pipeline, shown in Fig.~\ref{fig:pipeline}. 
Our inference mainly determines whether to repeat words to a case by measuring the voting consistency of video clips with the proposed Matching Entropy (ME) metric.
Additionally, we follow the standard inference process in method~\cite{Li2021Align}, which contains two steps: 1) There are several video candidates selected from the search video database using the video-caption feature similarity. 2) The correct video is obtained by ranking the video-caption matching.

\subsubsection{Voter Generation}
\ 
\newline

For text-to-video retrieval, we need to determine whether the video $V_i, V_i \in Database$ matches the caption $C_i$. In the first inference step, we use similarity metric to select $k$ video candidates $\{ V_1,V_2,...,V_k \}$. Then, we segment each video into $w$ short clips $\{v_0 | v_{1},v_{2},...,v_{w}\}$ using DDM-Net~\cite{Tang2022Progressive}, which is a generic event boundary detection method. $v_0$ represents the original video. After segmentation, we also merge some video clips with short duration. For videos with segments less than $w$, we use the original video for padding. Further, we employ the trained model to match the caption $C_i$ with $w+1$ video clips. For each video, we get $w+1$ matching scores $\{S_{0}, S_{1}, S_{2},..., S_{w}\}$. We calculate the matching scores for all $k$ video candidates and we get $k$ group scores (left):

\begin{equation}\label{Eq. 7}
S=\left[\begin{array}{cccc}
S^1_0 & S^1_1 & \ldots & S^1_w \\
S^2_0 & S^2_1 & \cdots & S^2_w \\
\vdots & \vdots & \ddots & \vdots \\
S^k_0 & S^k_2 & \cdots & S^k_w
\end{array}\right] \rightarrow
\left[\begin{array}{cccc}
S^1_0 & S^2_0 & \ldots & S^k_0 \\
S^1_1 & S^2_1 & \cdots & S^k_2 \\
\vdots & \vdots & \ddots & \vdots \\
S^1_w & S^2_w & \cdots & S^k_w
\end{array}\right]
\end{equation}

Because the caption is searched from scores $\{ S^1_w, S^2_w,..., S^k_w \}$, we further perform a transpose operation on the matrix (right). In the standard video-caption retrieval, the video located at the index of the maximum value of vector $\{ S^1_w, S^2_w,..., S^k_w \}$ is considered by the model to be the most matching video to the given caption. Therefore, for $w+1$ videos, we consider them as $w+1$ voters.

\subsubsection{Judgement with Matching Entropy}
\label{sec: judgement with ME}
\ 
\newline

Each voter makes a vote to determine which video index they think best corresponds to the provided caption.
The ideal outcome would be if all voters unanimously agree that the $i$-th video is the correct match for the caption, and their selection is accurate. However, our experiments reveal that the voters frequently make diverse choices, as illustrated by the various scenarios depicted in Fig.~\ref{fig:pipeline}. 
This phenomenon indicates that the model lacks enough representation of the given caption. Just like humans learn challenging language content, humans use the repeating strategy to assist the learning process. We also employ the ``Repetition'' strategy for these cases and find it useful to enhance the model's performance. 

To identify these cases in the test set, we propose the Matching Entropy, which can reflect the degree of the consistency of voters. We set the frequency list $\{\vartheta_{0},\vartheta_{1},...,\vartheta_{w}\}$ of $top-1$ answers for $w+1$ voters. Specifically, the duplicate element in the frequency list is deleted. The probability $p_{i}$ of frequency $\vartheta_{i}$ is denoted as Eq. ~\ref{Eq. 8}.

\begin{equation}\label{Eq. 8}
p_i=\frac{\vartheta_i}{w+1}
\end{equation}

The ME metric is defined with Eq. ~\ref{Eq. 9}.

\begin{equation}\label{Eq. 9}
\mathrm{M} E=E\left[-\log p_i\right]=-\sum_{i=1}^n p_i \log p_i
\end{equation}

The ME measures the degree of agreement among the voting results. The higher ME indicates a lower correspondence between the video and the caption. We set a threshold $\delta$ to filter the candidates that are not suitable for using the ``Repetition'' strategy.
When the ME value is larger than the $\delta$, we repeat the words to the original caption, which are the nouns and verbs in the caption. After identifying the cases in the test set that are suitable for ``Repetition'' strategy, we then conduct retrieval evaluation with new captions to obtain the final results. 

\section{Experiments}

\begin{table*}[ht]
    \centering
    \caption{
    Evaluations of text-to-video performance on MSR-VTT and DiDeMo benchmarks. We compare our method with three kinds of methods, including (a) based on pre-training on multimodal data, (b) based on pre-training on single-modal data, and (c) without pre-training. We mark the optimal method in bold and the sub-optimal method with an underline.}
    \label{tab: t2v results on MSR-VTT and DiDeMo}
    \small
    \begin{tabular}{lclclcrrcrccc}
       \toprule
       \multirow{2}{*}{\textbf{Methods}}  &&\multirow{2}{*}{\textbf{Pre-load Model}} &&\textbf{Pre-trained} && \multicolumn{3}{c}{\textbf{MSR-VTT}}  && \multicolumn{3}{c}{\textbf{DiDeMo}} \\
        \cmidrule{7-9} \cmidrule{11-13}
        && &&\textbf{Data} && R@1 & R@5 & R@10 && R@1 & R@5 & R@10 \\ \midrule
       \rowcolor{lightgray!20} \multicolumn{13}{l}{\textit{(a) Pre-training on Multimodal Data}}\\\
       \multirow{2}{*}{VALOR~\cite{chen2023valor}} && \multirow{2}{*}{CLIP} && Video (2.5M)+Image (3M) && \multirow{2}{*}{36.2} & \multirow{2}{*}{64.7} & \multirow{2}{*}{75.4} && \multirow{2}{*}{43.2} & \multirow{2}{*}{73.9} & \multirow{2}{*}{82.4} \\
        &&  && +Video-audio (1M) &&  &  &  &&  &  &  \\
       TemPVL~\cite{Ma2023Temporal}  && SwinT+ViT && Video (2.5M)+Image (3M) && 41.0 & 68.2 & 77.7 && 48.6 & 76.1 & \textbf{85.4} \\
       UMT-B~\cite{li2023umt-b}  && ViT-B/16 && Video (2M)+Image (3M) && 46.3 & 72.7 & 82.0 && - & - & - \\ 
       VINDLU-L~\cite{cheng2023vindlu} && ViT-B/16 && Video (10M)+Image (15M) && \underline{48.8} & 72.4 & 82.2 && - & - & - \\
       CLIP-ViP~\cite{xue2022clip-vip}  && CLIP && Video (102.5M)+Image (3M) && \textbf{49.1} & 73.1 & \textbf{83.5} && 47.0 & 75.3 & 84.1 \\
       \midrule
       \rowcolor{lightgray!20} \multicolumn{13}{l}{\textit{(b) Pre-training on Single-modal Data}}\\
       CLIP-ViP~\cite{xue2022clip-vip} && CLIP && Image (12.3M) && 45.6 & 70.7 & 81.1 && 43.7 & 69.5 & 77.9 \\
       CLIP4CLIP~\cite{Luo2022CLIP4Clip} && CLIP && Video (0.38M) && 43.5 & 70.7 & 80.5 && - & - & - \\ \midrule
       \rowcolor{lightgray!20} \multicolumn{13}{l}{\textbf{\textit{(c) Without Pre-training}}}\\
       CLIP4CLIP~\cite{Luo2022CLIP4Clip} && CLIP && No && 42.1 & 71.9 & 81.4 && 43.3 & 70.2 & 80.6 \\
       X-Pool~\cite{Gorti2022X-pool} && CLIP && No && 43.9 & 72.5 & 82.3 && - & - & - \\
       MuMUR~\cite{madasu2023mumur} && CLIP (ViT-B/32) && No && 44.8 & 72.0 & 82.5 && 44.4 & 74.3 & 83.1 \\
       BLIP~\cite{Li2022Blip} (fine-tuned) && BLIP && No && 44.9 & 68.7 & 77.9 && 45.2 & 73.1 & 82.9 \\
       ProST~\cite{li2023prost} && CLIP (ViT-B/16) && No && 46.9 & \textbf{73.3} & \underline{82.9} && 47.5 & 75.5 & 84.4 \\
       % TEFAL~\cite{ibrahimi2023tefal} && ViT-B/32 && No && 48.1 & \textbf{73.8} & 82.8 && - & - & - \\
       \cmidrule{1-1} \cmidrule{3-3} \cmidrule{5-5} \cmidrule{7-9} \cmidrule{11-13}
       Ours (w/o ``R'') && BLIP && No && 47.0 & 71.2 & 80.0 && \underline{49.6} & \underline{76.5} & \underline{84.6} \\
       \textbf{Ours (w/ ``R'')} && BLIP && No && \textbf{49.1} &\underline{73.2} & 81.2 && \textbf{51.2} & \textbf{76.9} & 84.1 \\
       \bottomrule
    \end{tabular}
\end{table*}

\subsection{Experimental Settings}
\label{experimental-settings}

\textbf{Evaluation benchmarks.}
We evaluate the performance of our proposed framework on four commonly-used video datasets, including MSR-VTT ~\cite{xu2016msrvtt}, DiDeMo ~\cite{Hendricks2017didemo}, MSVD~\cite{chen2011msvd}, and Youcook2~\cite{zhou2018youcook2}.
MSR-VTT dataset is a large-scale dataset for open-domain video retrieval and captioning. It contains 10,000 videos from 20 categories and each video has 20 captions to describe the contents of the video. Following~\cite{Gorti2022X-pool, Luo2022CLIP4Clip}, we conduct the experiments on MSR-VTT-7k, that 7,000 videos for training and 1,000 videos for testing.
DiDeMo~\cite{Hendricks2017didemo} dataset has 10,000 videos paired with 40,000 captions. It covers various human activities and consists of video clips under 30 seconds. Following works~\cite{Lei2021Less, Luo2022CLIP4Clip}, we concatenate all captions of one video to a single caption for retrieval evaluation.
YouCook2~\cite{zhou2018youcook2} dataset contains 2,000 videos from 89 cooking recipes and each recipe average has 22 videos.
MSVD~\cite{chen2011msvd} dataset has 1,970 videos with 80,000 captions. We use 1,200 videos for training and 670 videos for testing.

\noindent \textbf{Evaluation metrics.}
We use standard retrieval metrics $Recall@N$ at rank $N, N \in \{1,5,10\}$ on all the datasets for both the text-to-video and the video-to-text tasks. The metric represents the percentage of correct videos among the top $N$ retrieved videos or texts with given textual queries or visual queries.

\noindent \textbf{Implementation details.}
We initial the parameters of the proposed model from the BLIP model~\cite{Li2022Blip}.
To encode the vision parts, the visual transformer~\cite{Dosovitskiy2020An} and the spatial-temporal layer in~\cite{Bertasius2021Is} are employed. The video encoder learns spatio-temporal features from frame-level patches with divided space-time attention. This structure applies temporal and spatial attention sequentially and independently. The blocks of the spatial-temporal module are also initialized by reusing the BLIP model.
Besides, we choose the BERT~\cite{Devlin2019BERT} as the text coder, where a [CLS] token is also added.
We implement our framework in PyTorch and V100 Nvidia GPUs. The initial learning rate is set to 5e-6 for the whole model parameters.
During model training, eight frames are randomly sampled from videos.
\textbf{In all experiments, method ``w/ `R' '' denotes we use the proposed inference pipeline with ``Repetition'' strategy to test the model, and ``w/o `R' '' represents the direct inference.}

\subsection{State-of-the-Arts Retrieval Models}

\textbf{MSR-VTT and DiDeMo datasets.} We conduct a comprehensive comparison with existing works on the MSR-VTT and DiDeMo datasets in Table~\ref{tab: t2v results on MSR-VTT and DiDeMo}, including methods: (a) based on pre-training on large-scale multimodal data, \textit{i.e.}, the video and the image; (b) based on pre-training on single-modal data, \textit{i.e.}, the video or the image; (c) without the pre-training stage. Methods (a) and (b) generally employ a large amount of video or image data for pre-training~\cite{chen2023valor, Ma2023Temporal, li2023umt-b, cheng2023vindlu, xue2022clip-vip}, they are also fine-tuned on MSR-VTT and DiDeMo for inference. The ``pre-training and finetuning'' paradigm endows these models with a significant improvement on downstream datasets. However, compared to these methods, our model achieves better or more competitive results. For example, our model outperforms methods like UMT-B~\cite{li2023umt-b} and VINDLU-L~\cite{cheng2023vindlu}, which have been fine-tuned on large-scale datasets. Compared to VINDLU-L~\cite{cheng2023vindlu}, our model achieves an R@1 of 49.1, while VINDLU-L's score is 48.8. Especially the CLIP-ViP method~\cite{xue2022clip-vip}, which uses an amazing 102.5M video data for pre-training. In contrast, our method only requires the proposed inference pipeline with a ``Repetition'' strategy to achieve the same R@1 accuracy of 49.1.
Furthermore, compared to other similar methods~\cite{Luo2022CLIP4Clip, Gorti2022X-pool, madasu2023mumur, Li2022Blip, li2023prost, ibrahimi2023tefal} that have not undergone pre-training stage, our proposed method achieves the best average performance on these two datasets, especially the R@1 metric. Our model obtained an R@1 value of 49.1 on the MSR-VTT and an R@1 value of 51.2 on the DiDeMo.
So, it is noteworthy that our proposed ``Repetition'' strategy can significantly enhance the model's performance. Compared to not employing the ``Repetition'' strategy, the method utilizing it achieved an increase of 2.1 in R@1 on the MSR-VTT and 1.6 on the DiDeMo, respectively. \textit{Noticed that such improvement does not need extra data for training.}
Besides, with our proposed ``Repetition'' strategy, we also find that the effect of R@1 is improved the most compared with R@5 and R@10. 
Because we set the frequency list on the top-1 answers for all the voters in Section~\ref{sec: judgement with ME}, it will make the model more sensitive to the R@1.

\begin{table}[H]
\centering
\caption{Text-to-video performances on MSVD.}
\label{tab: t2v results on MSVD}
\small
\setlength{\tabcolsep}{4.2mm}{
\begin{tabular}{lccc}
\toprule
\textbf{Methods}                   & \textbf{R@1}  & \textbf{R@5}  & \textbf{R@10}  \\ \midrule
\rowcolor{lightgray!20} CenterCLIP~\cite{Zhao2022CenterCLIP}     & 50.6          & 80.3          & 88.4  \\
\rowcolor{lightgray!20} CAMoE~\cite{cheng2021CAMoE}     & 51.8          & 87.6          & 87.6  \\ \midrule
CLIP4CLIP~\cite{Luo2022CLIP4Clip}     & 46.2         & 76.1          & 84.6        \\
MuMUR~\cite{madasu2023mumur}       & 47.2          & 77.3          & 86.2              \\
X-Pool~\cite{Gorti2022X-pool}     & 47.2          & 77.4          & 86.0         \\
NCL~\cite{Park2022Normalized}     & 47.8          & 77.5          & 85.9           \\
DMAE~\cite{jiang2023dmae}     & 48.7          & 78.4          & 86.3         \\
\cmidrule(lr){1-1} \cmidrule(lr){2-2} \cmidrule(lr){3-3} \cmidrule(lr){4-4}
Ours (w/o ``R'')                     & 50.6          & 78.7          & 86.8           \\
\textbf{Ours (w/ ``R'')}             & \textbf{51.9}(+1.3)          & \textbf{79.6}(+0.9)          & \textbf{87.0}(+0.2)         \\ \bottomrule
\end{tabular}}
\end{table}

\noindent \textbf{MSVD datasets.}
In Table~\ref{tab: t2v results on MSVD}, we evaluate the text-to-video retrieval performance of our methods on the MSVD dataset. In the table, our method achieved an accuracy of 51.9 at R@1, 79.6 at R@5, and 87.0 at R@10, outperforming existing approaches. Our method with ``Repetition'' outperforms the without by 1.3\% at R@1, 0.9\% at R@5, and 0.2\% at R@10. Noticed that the methods with gray backgrounds are pre-training on more data to have better performances. Besides, it can be observed that our proposed ``Repetition'' strategy is also useful on the MSVD dataset, which can improve our framework's retrieval performance.

\begin{table}[H]
\centering
\caption{Text-to-video performances on YouCook2.}
\label{tab: t2v and v2t results on Youcook2}
\small
\setlength{\tabcolsep}{5.5mm}{
\begin{tabular}{lccc}
\toprule
\textbf{Methods}                   & \textbf{R@1}  & \textbf{R@5}  & \textbf{R@10}  \\ \midrule
RoME~\cite{Satar2022RoME}     & 6.3          & 16.9          & 25.2           \\
E-at-Once~\cite{Shvetsova2022everything}     & 13.7          & 35.3          & 48.4         \\
\cmidrule(lr){1-1} \cmidrule(lr){2-2} \cmidrule(lr){3-3} \cmidrule(lr){4-4}
\textbf{Ours (w/ ``R'')}    & \textbf{18.3}   & \textbf{39.2}    & \textbf{50.2}     \\ \bottomrule
\end{tabular}}
\end{table}

\noindent \textbf{YouCook2 dataset.} Additionally, YouCook2 dataset is also utilized to evaluate the retrieval performance in works~\cite{Satar2022RoME, Shvetsova2022everything}. We compare with these methods in Table~\ref{tab: t2v and v2t results on Youcook2}. It can be found that our method achieves the best performance.

\begin{table}[H]
\centering
\small
\caption{Video-to-text performances on MSR-VTT.}
\label{tab: results of v2t on msrvtt}
\setlength{\tabcolsep}{3.3mm}{
\begin{tabular}{lccc}
\toprule
\textbf{Methods}                   & \textbf{R@1}  & \textbf{R@5}  & \textbf{R@10}  \\ \midrule
CLIP4CLIP ~\cite{Luo2022CLIP4Clip}       & 43.1          & 70.5          & 81.2              \\
BLIP~\cite{Li2022Blip} (fine-tuned)     & 43.5          & 67.4          & 78.2           \\
NCL~\cite{Park2022Normalized}     & 44.9          & 71.8          & 80.7        \\
MuMUR~\cite{madasu2023mumur}     & 45.5          & 73.4          & \textbf{84.7}         \\
TS2-Net~\cite{Liu2022TS2-Net}  & 45.6          & 73.5          & 83.2         \\
CMA~\cite{Jiang2022Cross}      & 46.2          & \textbf{73.6}          & 83.8  \\
CenterCLIP~\cite{Zhao2022CenterCLIP}      & 46.9          & 73.4          & 83.2  \\ 
\cmidrule(lr){1-1} \cmidrule(lr){2-2} \cmidrule(lr){3-3} \cmidrule(lr){4-4}
Ours (w/o ``R'')                     & 46.2          & 71.6          & 80.1           \\
\textbf{Ours (w/ ``R'')}             & \textbf{48.2}          & 72.9          & 80.8         \\ \bottomrule
\end{tabular}}
\end{table}

\noindent \textbf{Video-to-text on MSR-VTT.}
To completely test the model's ability, we evaluate the video-to-text performance on the MSR-VTT dataset. 
Since many methods in Table~\ref{tab: t2v results on MSR-VTT and DiDeMo} do not report video-to-text values, besides models~\cite{Luo2022CLIP4Clip, Li2022Blip, madasu2023mumur}, we also compare with methods~\cite{Park2022Normalized, Liu2022TS2-Net, Jiang2022Cross, Zhao2022CenterCLIP}. Shown in Table~\ref{tab: results of v2t on msrvtt}, our method also demonstrates commendable performance in the video-to-text task. 
Although our method slightly lags behind in R@5 and R@10, it demonstrates outstanding performance in the R@1 metric, improving by 1.3 compared to CenterCLIP~\cite{Zhao2022CenterCLIP}. Moreover, the top-1 retrieval accuracy can have a significantly stronger impact in practical applications, which is why methods~\cite{Luo2022CLIP4Clip,Gorti2022X-pool, Li2022Blip} have received special attention.
The above performances indicate that our model provide a strong ability that can convert visual information in video content into verbal descriptions and match it from the search caption database.

\begin{table}[h]
\centering
\caption{Comparison with other fine-grained video retrieval methods in text-to-video performance.} 
\label{tab: compare with other fine-grained method}
\small
\setlength{\tabcolsep}{1.1mm}{
\begin{tabular}{lcccccc}
\toprule
\multirow{2}{*}{\textbf{Methods}}  & \multicolumn{3}{c}{\textbf{DiDeMo}}       & \multicolumn{3}{c}{\textbf{MSVD}}    \\ 
                 \cmidrule(lr){2-4} \cmidrule(lr){5-7}
& \textbf{R@1}    & \textbf{R@5}    & \textbf{R@10}    & \textbf{R@1} &    \textbf{R@5} &    \textbf{R@10} \\
\midrule
UCoFIA~\cite{wang2023unified}   & 46.5   & 74.8   & -    & 47.4     & 77.6   & -           \\
Align\&Tell~\cite{wang2022align}    &  -  &  -  &  -   & 49.3     & 79.1   & \textbf{87.9}           \\
X-CLIP~\cite{ma2022x-clip}  &  47.8  &  \textbf{79.3}  &  -   & 50.4     & \textbf{80.6}   & - \\
\cmidrule(lr){1-1} \cmidrule(lr){2-4} \cmidrule(lr){5-7}
\textbf{Ours (w/o ``R'')}   & \textbf{49.6}(+1.8) & 76.5 & \textbf{84.6} & \textbf{50.6}(+0.2) & 78.7 & 86.8  \\
\bottomrule
\end{tabular}}
\end{table}

\noindent \textbf{Comparison with Fine-grained Methods.} 
To demonstrate the effectiveness of the fine-grained data and objective, we compare our method with existing fine-grained methods~\cite{wang2023unified, wang2022align, ma2022x-clip} of text-to-video retrieval on DiDeMo and MSVD datasets, respectively.
For a fair comparison, we do not use the ``Repetition'' strategy. In Table~\ref{tab: compare with other fine-grained method}, it can be observed that our method outperforms existing methods. Regarding the R@1 metric, our method surpasses previous methods by 1.8 on the DiDeMo dataset and by 0.2 on the MSVD dataset. Although the X-CLIP method~\cite{ma2022x-clip} performs better on the R@5 metric, the evaluation for R@1 is more difficult, indicating that our model is superior in terms of precision compared to it.

\subsection{Ablation Study}

\begin{table*}[htb]
  \setlength{\tabcolsep}{2.4mm}
    \centering
    \caption{Comparison with different coarse-to-fine objectives in our framework. VCC: Video-Caption Contrastive objective. VCM: Video-Caption Matching objective. LM: Language Modeling objective. FTC: Frame-Title Contrastive objective. FTM: Frame-Title Matching objective. Coarse-grained objectives: VCC, VCM, and LM. Fine-grained objectives: FTC and FTM.} \vspace{-6pt}
    \label{tab: evaluation of different objectives}
    \small
    \begin{tabular}{lcrcrcrcrcrcrcrcrcr}
       \toprule
       \multirow{2}{*}{\textbf{Methods}} &&\multirow{2}{*}{\textbf{VCC}} &&\multirow{2}{*}{\textbf{VCM}} &&\multirow{2}{*}{\textbf{LM}} &&\multirow{2}{*}{\textbf{FTC}} &&\multirow{2}{*}{\textbf{FTM}} && \multicolumn{3}{c}{\textbf{Text-to-Video}} && \multicolumn{3}{c}{\textbf{Video-to-Text}} \\
        \cmidrule{13-15} \cmidrule{17-19}
        && && && && && && \textbf{R@1} & \textbf{R@5} &\textbf{R@10}  && \textbf{R@1} & \textbf{R@5} &\textbf{R@10} \\
       \midrule
        \multirow{4}{*}{Framework}&&\Checkmark &&\Checkmark && \XSolidBrush &&\XSolidBrush &&\XSolidBrush && 44.9 & 68.7 & 77.9 && 43.5 & 67.4 & 78.2 \\
        &&\Checkmark &&\Checkmark && \Checkmark &&\XSolidBrush &&\XSolidBrush && 45.9 & 70.7 & 79.2 && 42.6 & 68.1 & 75.7 \\
        &&\Checkmark &&\Checkmark &&\XSolidBrush &&\Checkmark &&\XSolidBrush && 46.2 & \textbf{71.2} & 79.9 && 44.4 & 71.3 & 79.3 \\
        \rowcolor{lightgray!20} &&\Checkmark &&\Checkmark &&\XSolidBrush &&\XSolidBrush &&\Checkmark  && \textbf{47.0} & \textbf{71.2} & 80.0 && \textbf{46.2} & \textbf{71.6} & 80.1 \\
        &&\Checkmark &&\Checkmark &&\XSolidBrush &&\Checkmark &&\Checkmark  && 46.5 & 71.0 & \textbf{80.2} && 45.0 & 71.3 & \textbf{80.2} \\
       \bottomrule
    \end{tabular}
\end{table*}

\noindent \textbf{Effectiveness of the proposed coarse-to-fine objectives.}
To obtain fine-grained features for better retrieval performance, we consider the contrastive and matching objectives between the vision and language data. Among these objectives, we explore their effectiveness in Table~\ref{tab: evaluation of different objectives}.
To maintain the basic power of the pre-trained model, we keep the VCC and the VCM objectives and use them as the baseline. Based on these two objectives, we design the following settings: (a) only contains LM; (b) only contains FTC; (c) only contains FTM; (d) contains FTC and FTM.
It can be found that the model with fine-grained objectives, \textit{e.g.,} the setting (b) or the setting (c), outperforms coarse-grained objectives, \textit{e.g.,} the baseline or the setting (a).
However, setting (a) shows an improvement in the text-to-video task but a decline in performance on the video-to-text task.
When using fine-grained video frame and title information, the model's performance is further improved, with FTM outperforming FTC. For example, in the text-to-video task, the R@1 metric is 47.0 vs. 46.2, in settings (c) and (b).
Additionally, We observe that when both FTC and FTM metrics are used simultaneously, the model's performance actually decreases. We speculate that the contrastive objective may lead to over-matching of specific negative pairs, resulting in suboptimal learning of the true relationship between video frames and text. By using only the matching objective, the model focuses on strengthening the correct alignment without overfitting to distractor information.

\begin{table}[h]
\caption{Effectiveness of ``Repetition'' strategy on other video-language retrieval models, we provide the text-to-video results on MSR-VTT and DiDeMo.} 
\label{tab: different model with repetition}
\small
\setlength{\tabcolsep}{1.6mm}{
\begin{tabular}{lcccccc}
\toprule
\multirow{2}{*}{\textbf{Methods}}    & \multicolumn{3}{c}{\textbf{MSR-VTT}}       & \multicolumn{3}{c}{\textbf{DiDeMo}}         \\
                 \cmidrule(lr){2-4} \cmidrule(lr){5-7}
 & \textbf{R@1}  & \textbf{R@5}  & \textbf{R@10} & \textbf{R@1} & \textbf{R@5} & \textbf{R@10} \\ \midrule
ALPRO (w/o ``R'')       & 33.9   & 60.7   & \textbf{73.2}    & 35.9     & 67.5   & 78.8           \\
\textbf{ALPRO (w/ ``R'')}         & \textbf{34.8}  & \textbf{61.3}   & 72.8   & \textbf{36.6}     &  \textbf{67.8}  & \textbf{78.9}   \\
\midrule
BLIP (w/o ``R'')        & 44.9          & 68.7          & 77.9          & 45.6          & 73.9          & \textbf{82.9}           \\
\textbf{BLIP (w/ ``R'')}  & \textbf{47.5} & \textbf{69.8} & \textbf{80.2} & \textbf{46.6} & \textbf{74.2} & 81.4  \\ \bottomrule
\end{tabular}}
\centering
\end{table}

\noindent \textbf{Effectiveness of ``Repetition'' strategy on other models.} 
To validate the robustness of the proposed inference pipeline with ``Repetition'', we apply the ``Repetition'' strategy on two representative video-language retrieval models, \textit{i.e.,} the ALPRO~\cite{Li2021Alignandprompt} and the BLIP~\cite{Li2022Blip}. We fine-tune the two models on MSR-VTT and DiDeMo, respectively. The learning rate is set as $1e-5$ and the batch size is set as 48.
Surprisingly, we find that our proposed inference pipeline with the ``Repetition'' strategy is effective for both models.
In Table~\ref{tab: different model with repetition}, compared with ``BLIP (w/o `Repetition' )'' and ``BLIP (w/ `Repetition' )'', the latter outperforms the former by 2.6\% at R@1, 1.1\% at R@5, and 2.3\% at R@10.
We analyze that the proposed inference strategy is inspired by the human language learning process, which could potentially become an important method for improving the performance of vision-language models.

\begin{table}[h]
\centering
\caption{Ablation of adding the ``Repetition'' information to all cases in the test set or adding them only to the cases selected by our inference pipeline. We provide the results of text-to-video on MSR-VTT.}
\label{tab: impact of inference pipeline}
\small
\setlength{\tabcolsep}{4.0mm}{
\begin{tabular}{lccc}
\toprule
\textbf{Methods}                    & \textbf{R@1}      & \textbf{R@5}      & \textbf{R@10} \\ \midrule
\rowcolor{lightgray!20} Baseline    & 47.0              & 71.2              & 80.0          \\ 
Repetition for all                 & 45.2 $\downarrow$ & 70.9 $\downarrow$ & 79.4 $\downarrow$ \\ 
% ``R w/ All cases (twice)''          & 46.1 $\downarrow$ & 71.4              & 80.0          \\ 
\cmidrule(lr){1-1} \cmidrule(lr){2-2} \cmidrule(lr){3-3} \cmidrule(lr){4-4}
\textbf{Repetition for target}                & \textbf{49.1}$\uparrow$     & \textbf{73.2}$\uparrow$     & \textbf{81.2}$\uparrow$  \\ 
\bottomrule
\end{tabular}}
\end{table}

\noindent \textbf{Ablation of selecting target cases for ``Repetition''.}
The proposed inference pipeline with the ``Repetition'' strategy involves repeating words for some of the cases in the test set. However, when deciding where to add the words, the most naive approach is to add the repeated words to all cases. To explore whether it is necessary to develop a specially designed inference pipeline that uses a voting mechanism and matching entropy to select target cases, we conduct the following experiments: (a) Baseline: direct inference without using ``Repetition''; (b) ``Repetition for all'': adding repeated words to all cases in the test set; (c) ``Repetition for target'': adding repeated words only to the cases selected by the proposed inference pipeline.
Table~\ref{tab: impact of inference pipeline} shows that simply repeating words for all cases not only fails to improve the model's performance but also harms the model's effectiveness. We analyze this phenomenon and conclude that adding extra words to all cases introduces irrelevant and disruptive information into the original captions, which negatively impacts the matching ability, particularly the R@1 metric. Therefore, it is necessary to apply the ``Repetition'' strategy to specific cases via our proposed inference pipeline.

\begin{table}[h]
\centering
\caption{Effectiveness of the ME metric in inference pipeline. We evaluate the text-to-video on the MSR-VTT.}
\label{tab: impact of ME metirc}
\small
\setlength{\tabcolsep}{3.0mm}{
\begin{tabular}{lccc}
\toprule
\textbf{Methods}                                 & \textbf{R@1} & \textbf{R@5} & \textbf{R@10} \\ \midrule
\rowcolor{lightgray!20} Baseline                     & 47.0         & 71.2         & 80.0          \\ 
\cmidrule(lr){1-1} \cmidrule(lr){2-2} \cmidrule(lr){3-3} \cmidrule(lr){4-4}
Inference w/o ME             & 47.5 (+0.5)           & 72.0 (+0.8)           & 80.3 (+0.3)          \\
\textbf{Inference w/ ME}     & \textbf{49.1} (+2.1)  & \textbf{73.2} (+2.0)  & \textbf{81.2} (+1.2)  \\ 
\bottomrule
\end{tabular}}
\end{table}

\begin{figure*}[th]
\begin{center}
\includegraphics[width=1.0\linewidth]{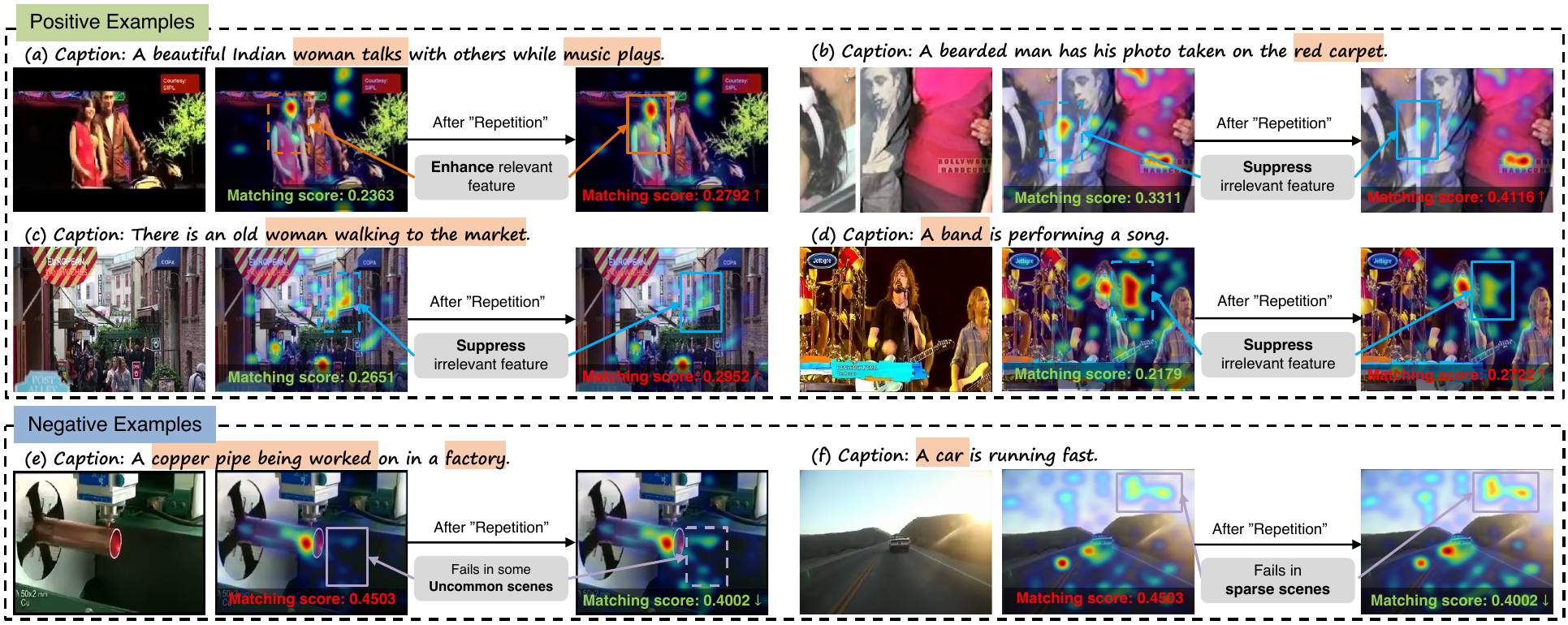}
\end{center}
   \caption{Visualizations of feature map after adding the ``Repetition'' for positive and negative examples.}
\label{fig: feature map visualization}
\vspace{-5pt}
\end{figure*}

\noindent \textbf{Impact of the proposed ME metric.} 
To validate the role of the ME metric in the inference pipeline, we conduct the experiments of text-to-video on MSR-VTT in Table~\ref{tab: impact of ME metirc}. The ``Baseline'' denote the direct retrieval inference without using the ``Repetition'' strategy, the ``Inference w/o ME'' represents that we use the voting results without the ME to select the target cases, and the ``Inference w/ ME'' is the complete method with ME. It can be observed that the method with ME significantly improves performance, which proves the effectiveness of the ME metric.

\begin{table}[t!]
\centering
\caption{Ablation study of different types of words for repetition in text-to-video on MSR-VTT and DiDeMo. The white background indicates words added to the beginning and the gray background to the end. We mark the optimal results in bold and the suboptimal with underline.} 
\label{tab: different words for repeating}
\small
\setlength{\tabcolsep}{2.0mm}{
\begin{tabular}{lcccccc}
\toprule
                 & \multicolumn{3}{c}{\textbf{MSR-VTT}}       & \multicolumn{3}{c}{\textbf{DiDeMo}}         \\
                 \cmidrule(lr){2-4} \cmidrule(lr){5-7}
\textbf{Methods} & \textbf{R@1}  & \textbf{R@5}  & \textbf{R@10} & \textbf{R@1} & \textbf{R@5} & \textbf{R@10} \\ \midrule
Baseline      & \underline{47.9}          & \underline{71.2}          & \underline{80.0}          & \underline{49.6}          & \underline{76.5}          & \textbf{84.6}           \\ \midrule
$\rm Noun^b$         & 41.4          & 68.2          & 77.8          & 43.8          & 71.9          & 80.3           \\
$\rm Verb^b$         & 43.1           & 68.3           & 77.9           & 47.3          & 72.8          & 81.7           \\
$\rm Noun+Verb^b$        & 38.2           & 63.7           & 74.8           & 40.1          & 68.0          & 77.3           \\ \midrule
\rowcolor{lightgray!20}
$\rm Noun^e$       & 42.0           & 69.7           & 79.3          & 46.2          & 73.2          & 81.1           \\
\rowcolor{lightgray!20}
$\rm Verb^e$         & 43.5           & 70.4           & 77.5           & 48.3          & 74.4          & 83.5    \\
\rowcolor{lightgray!20}
$\rm Noun+Verb^e$        & 41.8           & 66.6           & 77.7           & 45.1          & 72.2          & 79.8           \\ \midrule
\textbf{Ours}  & \textbf{49.1} & \textbf{73.2} & \textbf{81.2} & \textbf{51.2} & \textbf{76.9} & \underline{84.1}  \\ \bottomrule
\end{tabular}}
\end{table}

\noindent \textbf{Effectiveness of different types of words for repeating.}
To explore the impact of adding extra words to a sentence on alignment, we conduct the following experiment: We randomly select some random words from the most frequent 3,500 words in the USA Statistics~\cite{white1990growth, hsu2011vocabulary}, including nouns, verbs, and combinations of both, and add them uniformly to the cases that were selected through the inference pipeline. At the same time, we consider two methods of addition: one by adding the words at the beginning of the original sentence, labeled as $*^b$, and the other by adding them at the end of the original sentence, labeled as $*^e$. We provide the results in Table~\ref{tab: different words for repeating}, which indicate that: 1) Compared to the baseline, repeating random words in the caption decreases the model's performance; 2) The position where these words are inserted significantly influences the outcome. Specifically, adding the same words at the end of the caption yields better results than adding them at the beginning; 3) Inserting a verb improves the caption more than adding a noun. Using only nouns as repeating words significantly degrades the model's accuracy; 4) Only by adding content related to the original sentence, by directly selecting from the original sentence, can the model's performance be improved.

\subsection{Visualization}
To further explore the effectiveness of the ``Repetition'' strategy, we visualize the feature map of several examples in Fig.~\ref{fig: feature map visualization}.
In each case, the first column shows the video frame, the second column presents the feature activation map without using the ``Repetition'' inference pipeline, and the last column shows the corresponding results with the ``Repetition''.
We also marks the change of matching score between the caption and the video in the images.

\section{Conclusion}
We proposed a coarse-to-fine framework for video-language retrieval. We employed the coarse-grained and the fine-grained objectives for different modalities. We also observed the ``Repetition'' phenomenon, which repeats the keywords of captions for retrieval. This strategy can effectively improve the performance of video retrieval. To fit such ability to our model, we propose an inference pipeline to select the target cases for the ``Repetition'' strategy. Experimental results on four benchmarks demonstrate that the proposed method outperforms previous approaches and the effectiveness of the inference pipeline with ``Repetition'' strategy.

\bibliographystyle{ACM-Reference-Format}
\bibliography{sample-base}

\end{document}